\documentclass[conference]{IEEEtran}
\usepackage{cite}
\usepackage{amsmath,amssymb,amsfonts}
\usepackage{algorithmic}
\usepackage{graphicx}
\usepackage{textcomp}
\usepackage{xcolor}
\usepackage{booktabs}
\usepackage{hyperref}
\usepackage{svg}
\begin{document}

\title{Transformer-Guided Swarm Intelligence for Frugal Neural Architecture Search}

\author{\IEEEauthorblockN{Romain Amigon}
\IEEEauthorblockA{
\textit{Université du Québec à Chicoutimi (UQAC)}\\
Saguenay, Canada \\
Email: amigonromaina@gmail.com / ramigon@etu.uqac.ca}
}

\maketitle

\begin{abstract}
Neural Architecture Search (NAS) has automated the design of deep learning models but traditionally requires massive computational resources, often measured in thousands of GPU-days. In this paper, we propose a frugal and memetic NAS framework designed to democratize architecture design on consumer-grade hardware. Our approach combines the global macro-search capabilities of an autoregressive Transformer controller, trained via Reinforcement Learning (RL), with the local micro-exploitation of an Artificial Bee Colony (ABC) algorithm. To prevent premature convergence during the RL phase, we introduce a dynamic entropy mechanism that forces topological exploration upon detection of performance stagnation. Evaluated on a standard GPU (NVIDIA RTX 3060), our hybrid method effectively resolves the "cold-start" problem inherent in metaheuristics. By algorithmically penalizing network depth, our framework actively mitigates model bloat: on the CIFAR-10 dataset, it discovers an efficient architecture reaching 84.85\% accuracy with only \textbf{$\sim$174,000 parameters}—significantly smaller than standard baselines like ResNet-20—in 3 hours of search time. Furthermore, we demonstrate the framework's flexibility by applying it to credit card fraud detection, directly optimizing the F1-Score on highly imbalanced tabular data to reach a F1-Score of 0.71 with a compact network of $\sim$4,600 parameters. These results suggest that our approach can yield tailored, accessible, and highly parameter-efficient deep learning models suitable for edge deployment.
\end{abstract}


\section{Introduction}
The design of neural networks has historically been dominated by an empirical approach, where hyperparameters and topology are chosen based on human experience. While Neural Architecture Search (NAS) has formalized this process by automating the discovery of optimal topologies, early Reinforcement Learning (RL) based methods faced prohibitive costs. Foundational works required hundreds of GPUs running for weeks \cite{zoph2017}.

Today, the field of NAS faces two intertwined challenges. The first is over-parameterization: the literature tends to produce architectures that are computationally redundant relative to the intrinsic complexity of their task. The second is a methodological gap in optimization. Traditional NAS controllers often rely on Recurrent Neural Networks (RNNs or LSTMs), which suffer from an information bottleneck. In a neural network, the choice of a specific layer strongly dictates the necessity of subsequent layers. LSTMs struggle to capture these long-range topological dependencies. Furthermore, while RL excels at finding a global macro-architecture, it is notoriously inefficient at fine-tuning continuous, micro-level hyperparameters.

This paper presents \textit{nas-torch}, a highly modular, "white-box" NAS framework specifically designed for applied engineering and resource-constrained environments. This work does not target state-of-the-art accuracy on large benchmarks, but rather serves as a proof of concept to demonstrate that highly efficient, task-specific feature extractors can be generated autonomously on hardware as constrained as a single consumer GPU. Rather than targeting marginal accuracy gains on massive compute clusters, our approach prioritizes democratic accessibility and structural flexibility. To this end, \textit{nas-torch} provides a suite of plug-and-play frugal optimizers—including Random Search, Simulated Annealing, Artificial Bee Colony (ABC), and an autoregressive Transformer—while allowing researchers to easily integrate custom search algorithms. 

The primary objective of this paper is to benchmark these diverse optimization strategies under strict computational budgets and demonstrate how algorithmic hybridization can resolve their inherent individual limitations. For instance, while pure swarm metaheuristics struggle with the "cold-start" problem in vast topological spaces, and Reinforcement Learning controllers lack micro-tuning granularity, we show that fusing them into a memetic pipeline (Transformer-guided ABC) effectively mitigates these bottlenecks. Ultimately, the framework completes this hybridized, highly efficient search in 3 hours on standard consumer-grade hardware.

Our main contributions include:
\begin{enumerate}
    \item We replace the RNN controller with a lightweight autoregressive Transformer controller containing $\sim$68,000 parameters, which organically leverages causal sequence representations to integrate smoothly as a structural prior into the memetic pipeline.
    \item A multi-objective reward function integrating a strict depth penalty to actively penalize model bloat for embedded and edge deployment.
    \item A memetic optimization pipeline bridging the macro/micro gap (Transformer "warm-start" followed by local ABC micro-exploitation) that significantly reduces the computational search cost compared to historical RL frameworks.
    \item The demonstration of the framework's domain agnosticism and practical flexibility, successfully optimizing tabular regression, medical classification, and highly imbalanced industrial fraud data on a single GPU.
\end{enumerate}

The source code is publicly available at: \url{https://github.com/Romain-Amigon/nas-torch}

\section{Related Work}
\textit{Reinforcement Learning NAS:} The approach pioneered by Zoph \& Le used Recurrent Neural Networks updated via Policy Gradient to generate architectures from scratch \cite{zoph2017}. While groundbreaking, these fundamental methods suffer from severe computational overhead (up to 22,400 GPU-days), isolating NAS research within well-funded industrial laboratories.

\textit{NAS Acceleration:} To mitigate this cost, structural shortcuts like Weight Sharing (ENAS) \cite{pham2018} or differentiable continuous spaces (DARTS) \cite{liu2019} were introduced, bringing search times down to a few GPU-days. However, continuous relaxation methods like DARTS can be unstable, sometimes suffering from performance collapse by over-selecting skip-connections. Furthermore, these methods are often tightly coupled to specific cell-based templates, lacking the general-purpose flexibility required to process non-vision modalities without manual restructuring.

\textit{Transformers in NAS:} More recently, the state-of-the-art has shifted towards utilizing Large Language Models (LLMs) to guide architecture search via code generation (e.g., GPT-NAS \cite{yu2025}). However, free-form LLM editors suffer from "Functional Entanglement"—inadvertently breaking topological constraints by modifying operators and wiring simultaneously. Such entanglement can be mitigated by constraining the LLM editing process, but existing approaches rely on heavy external models (e.g., DeepSeek-R1) acting as code editors. Our work positions itself differently: rather than relying on heavy LLMs that require external scaffolding, we build a compact, mathematically constrained Transformer generating discrete tokens. Paired with a swarm intelligence metaheuristic, this ensures topological validity and rapid convergence on consumer-grade hardware.

\textit{Swarm Intelligence and Metaheuristics:} Beyond gradient-based and evolutionary methods, Swarm Intelligence algorithms—such as Particle Swarm Optimization (PSO) and Ant Colony Optimization (ACO)—have been leveraged to navigate discrete NAS spaces \cite{lankford2024}. While these swarm methods excel at local exploitation, they inherently suffer from the "cold-start" problem: the swarm is typically initialized randomly, wasting computational budgets evaluating sub-optimal architectures. Our memetic framework explicitly addresses this bottleneck. We utilize the Transformer as a topological prior to "warm-start" the ABC algorithm, combining global structural intelligence with fine-grained local exploitation in a single pipeline.

\textit{Synthesis and Positioning:} Existing NAS paradigms often force applied engineers into compromises: RL is computationally prohibitive \cite{zoph2017}, continuous relaxations are domain-rigid \cite{liu2019}, modern LLM controllers are over-parameterized, and pure Swarm methods suffer from cold-start inefficiencies \cite{lankford2024}. By using a lightweight Transformer to provide a topologically sound warm-start to a swarm metaheuristic, \textit{nas-torch} offers a fast, stable, and frugal alternative tailored for accessible AI development.

\section{Proposed Methodology: Autoregressive Topology Generation}
\subsection{Problem Formulation and Search Space}
Traditionally, the design of neural network architectures is a manual process followed by weight optimization. Mathematically, this involves defining a function $g$ that maps a weight space to a function space $\mathcal{F}$, which is the inference in neural network, where the objective is to optimize the weights $w \in \mathbb{R}^n$ for a given training set:

\begin{equation}
g : \mathbb{R}^n \rightarrow \mathcal{F}
\end{equation}

Our framework introduces a higher level of abstraction by defining a mapping function $f$ that accepts a specific architectural configuration as input and returns the corresponding function $g$. 

An architecture is represented by its topology, modeled as a graph $A$, and its layer-specific hyperparameters $X$. To facilitate the search process, we encode these dimensions into a unified representation matrix $\Theta$. Consequently, the mapping is defined as:

\begin{equation}
f : \Theta \rightarrow \mathcal{F}
\end{equation}
\begin{equation}
f(\theta) = g_{\theta}
\end{equation}

The primary objective of this research is to develop and evaluate a Transformer optimizer trained by reinforcement learning that maximizes the performance of the function $f$ over a validation dataset $\mathcal{D}_{val}$:

\begin{equation}
\theta^* = \text{arg}\max_{\theta \in \Theta} \text{Perf}(f(\theta), \mathcal{D}_{val})
\end{equation}

\begin{figure}[htbp] 
    \centering
    \includegraphics[width=\linewidth]{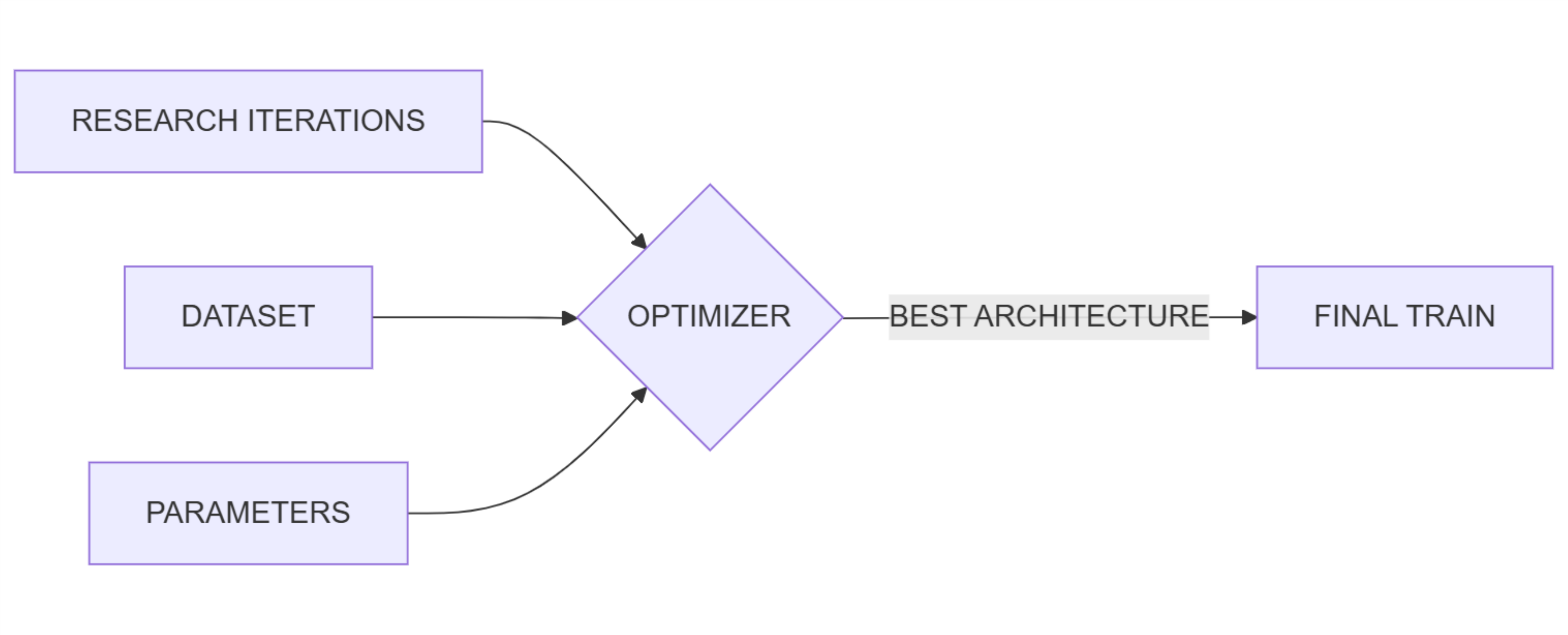}
    \caption{Overview state diagram of nas-torch}
    \label{fig:overview_NAS}
\end{figure}

\subsection{Autoregressive Transformer Controller}
Inspired by the foundational work on Neural Architecture Search \cite{zoph2017}, we use a controller network to generate the architectural hyperparameters of neural networks as a variable-length sequence. However, to better capture long-range topological dependencies, we replace the traditional Recurrent Neural Network (RNN) with a Transformer controller. 

We frame the generation process as a Natural Language Processing (NLP) task. The search space is discretized into a vocabulary of tokens (e.g., \texttt{conv\_3\_16}, \texttt{resblock\_32}, \texttt{flatten}). The Transformer, utilizing positional encoding and a causal mask, generates the topology sequentially.

The list of tokens predicted by the controller can be viewed as a sequence of actions $a_{1:T}$ designed to construct a child network. At the end of the proxy training phase, this child network achieves an accuracy $R$ on the held-out validation set. We use this performance metric as the reward signal to train the controller via reinforcement learning. More concretely, we ask our controller to maximize its expected reward, represented by $J(\theta_c)$:

\begin{equation}
J(\theta_c) = E_{P(a_{1:T};\theta_c)}[R]
\end{equation}

Since the reward signal $R$ is non-differentiable, we use the REINFORCE policy gradient method. To reduce high variance, we introduce a baseline term $b$, calculated as the moving average of previous architecture scores:

\begin{equation}
\nabla_{\theta_c} J(\theta_c) = \sum_{t=1}^{T} \mathbb{E}_{P} \left[ \nabla_{\theta_c} \log P(a_t | a_{1:(t-1)}; \theta_c) (R - b) \right]
\end{equation}

Crucially, to combat model bloat and force the controller to favor frugal architectures, the raw accuracy is not used directly. Instead, we compute a multi-objective reward incorporating a length penalty: $Reward = Accuracy - (\lambda \times Depth)$. The gradient update thus intrinsically teaches the Transformer to find the optimal trade-off between predictive power and architectural simplicity.

The continuous architecture space is discretized into a finite vocabulary of functional tokens. Specifically, our framework utilizes a strict dictionary of 15 precise tokens:
\begin{itemize}
    \item \textbf{Convolutions:} \texttt{conv\_3\_16}, \texttt{conv\_3\_32}, \texttt{conv\_5\_16} (denoting kernel size and output channels).
    \item \textbf{Residual Blocks:} \texttt{resblock\_16}, \texttt{resblock\_32}.
    \item \textbf{Pooling:} \texttt{pool\_2} (Max Pooling) and \texttt{avgpool} (Global Average Pooling).
    \item \textbf{Dense Layers:} \texttt{linear\_32}, \texttt{linear\_64}.
    \item \textbf{Regularization:} \texttt{dropout\_0.2}, \texttt{dropout\_0.5}.
    \item \textbf{Normalization:} \texttt{bn2d} (spatial) and \texttt{bn1d} (flattened context).
    \item \textbf{Structural:} \texttt{flatten} and the terminal \texttt{stop} token.
\end{itemize}
The Transformer, utilizing positional encoding and a causal mask, selects from this vocabulary to generate the topology sequentially until the \texttt{stop} token is predicted or the maximum depth is reached.

\subsection{Dynamic Entropy for Exploration}
To prevent premature convergence, the entropy of the probability distribution is integrated into the loss: $Loss = (- \log(P) \times Advantage) - (\beta \times Entropy)$. We implement a \texttt{variable\_entropy} function: if the score stagnates for $N$ iterations, $\beta$ increases, forcing the Transformer to generate divergent sequences to escape local minima.

\subsection{Proxy Evaluation Strategy}
Training every generated architecture to full convergence is computationally expensive. Therefore, we utilize a low-fidelity proxy evaluation strategy. 

During the search, each candidate is evaluated on a low-fidelity proxy built from the training set only: 40\% of the training images are used to train the candidate for a small number of epochs (typically 10), and a disjoint 10\% split is held out to score it. The official test set is never accessed during the search and is used only once, to evaluate the final architecture after full training. The final score assigned to the architecture is its best accuracy (or F1-Score) achieved on the validation split automatically generated by the framework from the train dataset given. 
To validate the reliability of this low-fidelity proxy, we conducted a rank correlation analysis. We sampled 15 architectures generated by the ABC optimizer and trained them for both 10 epochs (proxy) and 100 epochs (full training). The Spearman's rank correlation coefficient ($\rho$) between the proxy scores and the final accuracies is \textbf{0.721} ($p\text{-value} = 0.0024$). This statistically significant correlation confirms that the proxy provides a robust ranking signal, justifying its use to guide the search process efficiently.
It is important to note that these 15 architectures were sampled from the ABC optimizer's trajectory rather than a uniform random distribution. While this introduces a selection bias towards higher-performing regions, it accurately reflects the proxy's reliability within the actual subspace explored during the memetic search.

To further ensure frugality, a strict Early Stopping mechanism (\textit{patience}) is implemented: if the validation accuracy stagnates, the training is preemptively halted. The framework also dynamically infers the nature of the task from the dataset labels to assign the appropriate loss function (\texttt{CrossEntropyLoss} for multiclass or \texttt{BCEWithLogitsLoss} for binary tasks). Finally, if a generated architecture is topologically invalid, the evaluation method safely intercepts the exception and returns a score of $-\infty$, penalizing the genome without crashing the overall search process.
\begin{figure}[htbp] 
    \centering
    \includegraphics[width=\linewidth]{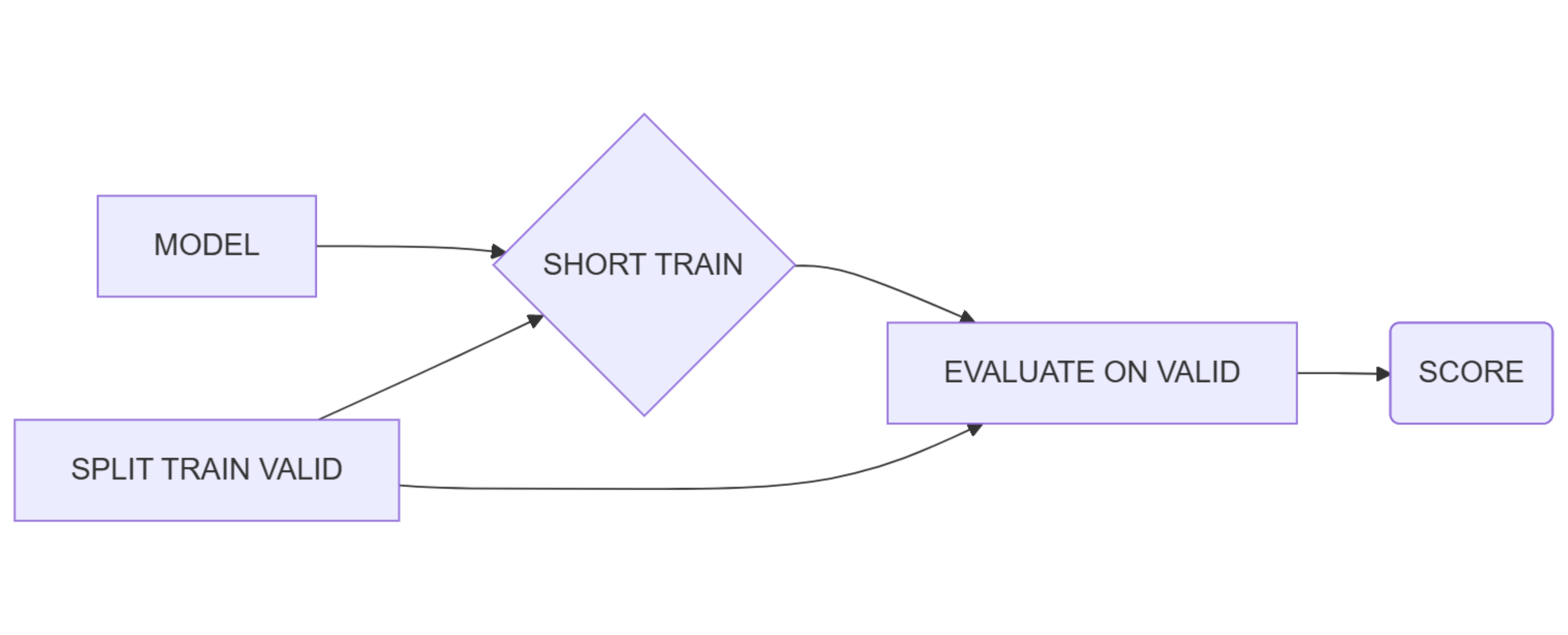}
    \caption{Proxy Evaluation Pipeline for each architecture}
    \label{fig:evaluate_pipeline}
\end{figure}

\begin{figure}[htbp] 
    \centering
    \includegraphics[width=\linewidth, height=0.85\textheight, keepaspectratio]{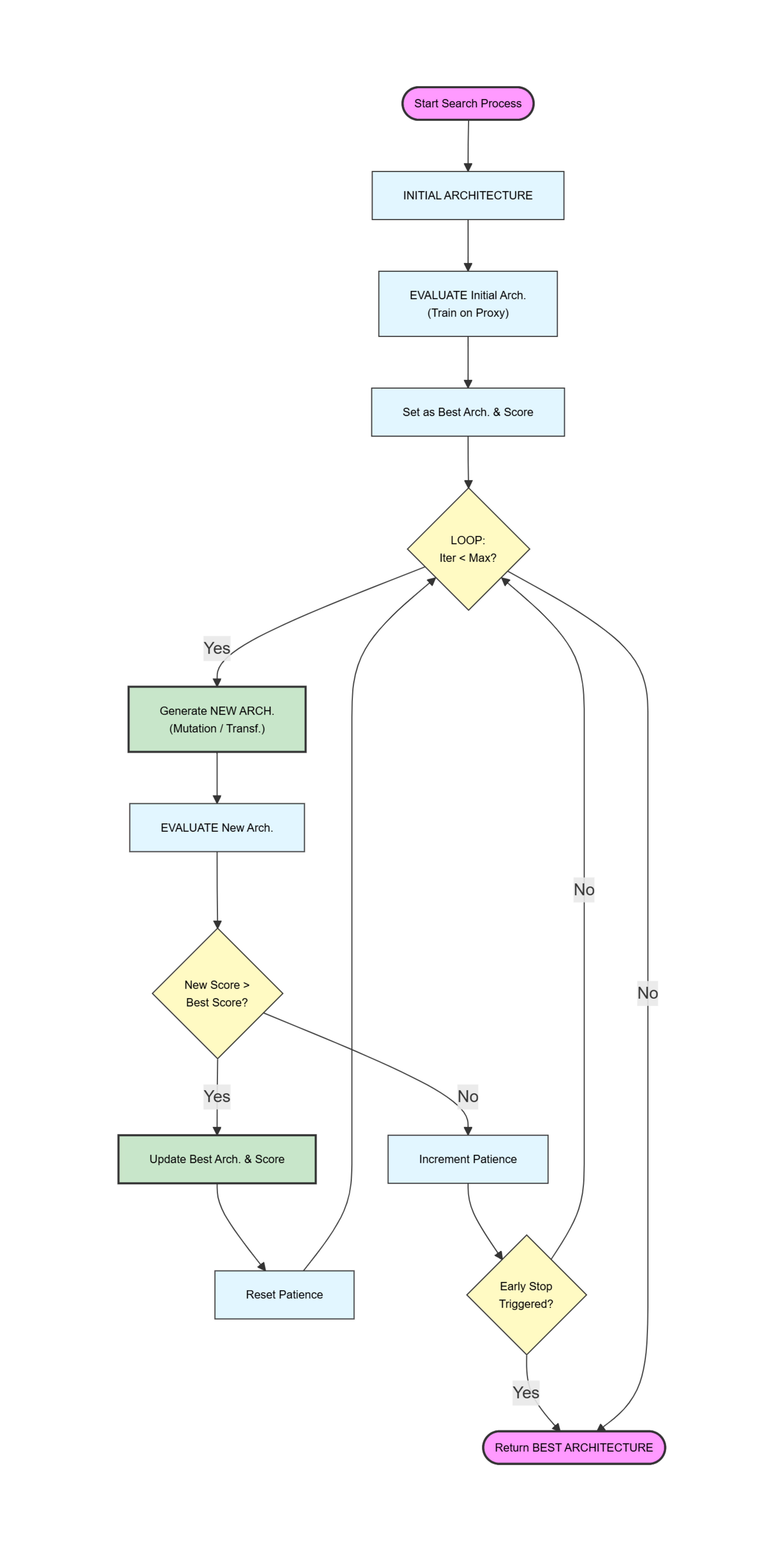}
    \caption{Global Optimizer Pipeline for each optimizer}
    \label{fig:pipeline_optimizer}
\end{figure}

\section{Proposed Methodology: Memetic Swarm Exploitation}

To compare our Transformer Optimizer, we developed alternative optimizers based on metaheuristics : Simulated Annealing, Genetic Algorithm and mainly, Artificial Bee Colony.

\subsection{Topological Consistency and Dynamic Parsing}
A major challenge in discrete NAS is ensuring that randomly mutated architectures remain mathematically valid. Modifying layers often leads to spatial dimension mismatches (e.g., connecting a convolutional layer directly to a linear layer without flattening). To address this, we developed a dynamic parser named \texttt{DynamicNet}. 

Before instantiating the PyTorch model, a topological reconciliation algorithm (\texttt{\_reconnect\_layers}) simulates a forward pass using a dummy tensor featuring the exact input shape of the dataset. As the tensor traverses the proposed computational graph, the parser automatically infers and injects the correct spatial and channel dimensions into the subsequent layers. This mechanism guarantees that any generated architecture will be natively evaluable, effectively eliminating compilation crashes during the automated search.

\subsection{Topological Mutation Operator}
To navigate the discrete search space, we designed a robust topological mutation operator (\texttt{neighbor}). When modifying an architecture, this operator randomly selects between four actions: altering continuous hyperparameters (e.g., modifying kernel sizes or channel depths), swapping activation functions, adding a layer, or removing a layer. To maintain coherence, a contextual verification mechanism (\texttt{is\_linear\_context\_check}) ensures that spatial operations (such as 2D Convolutions) are never inserted into flattened or linear spaces.

\section{Memetic Fine-Tuning via Artificial Bee Colony}
While the Transformer controller excels at extracting robust global macro-structures, it lacks the granularity to effectively fine-tune continuous hyperparameters. During preliminary experiments comparing Simulated Annealing (SA), Genetic Algorithms (GA), and the Artificial Bee Colony (ABC) algorithm, ABC demonstrated the highest stability and performance for local exploitation. 

The architecture generated by the Transformer is injected as a "Warm-Start" food source into the ABC metaheuristic. Employed and onlooker bees systematically explore the topological neighborhood using the aforementioned \texttt{neighbor} operator. Crucially, the ABC algorithm features a \texttt{limit} mechanism acting as a strict anti-stagnation protocol: if exploring the neighborhood of a specific architecture yields no improvement after a predefined number of trials, the scout bees abandon that region of the search space and reinitialize the food source. This prevents the swarm from being trapped in local minima and extracts the final percentages of performance from the Transformer's blueprint.

\section{Experiments and Results}

\subsection{Controller Architecture and Experimental Setup}
Designed specifically for frugality, the Transformer controller operates on a discrete search space mapped to a vocabulary of 15 functional tokens. 

The model embeds these tokens into a continuous representation of dimension $d_{model} = 64$, augmented by sinusoidal positional encodings. The core sequence modeling is handled by a 2-layer Transformer encoder, utilizing 4 parallel attention heads (\textit{nhead} = 4) and a feedforward network dimension of 128 ($d_{model} \times 2$). Regularization is applied via a dropout rate of 0.1. Optimized using Adam with a learning rate of 0.01, this controller contains approximately 68,000 trainable parameters. This minimal footprint ensures that the reinforcement learning policy updates execute in milliseconds, allowing the framework to run seamlessly on a constrained mobile GPU (NVIDIA GeForce RTX 3060 6GB) without memory bottlenecks.

\subsection{Domain Agnosticism and Optimization Baselines}
To validate the mathematical stability of our framework across diverse data modalities, we evaluated the optimizers on standard baseline datasets before tackling complex computer vision tasks. These included tabular regression (\textit{California Housing}) and binary medical classification (\textit{Breast Cancer}). We compared the generative controllers against traditional metaheuristics (Simulated Annealing and Artificial Bee Colony).

\begin{table}[htbp]
\caption{Baseline Performance on Tabular Datasets}
\begin{center}
\begin{tabular}{llc}
\toprule
\textbf{Task} & \textbf{Algorithm} & \textbf{Test Score (Avg $\pm$ Std)} \\
\midrule
\textbf{California Housing} & Simulated Annealing & -0.40 $\pm$ 0.03 (Neg MSE) \\
(Regression) & ABC Algorithm & \textbf{-0.32 $\pm$ 0.01} (Neg MSE) \\
& LSTM Controller (RL) & -0.39 $\pm$ 0.01 (Neg MSE) \\
& Transformer Controller & -0.36 $\pm$ 0.04 (Neg MSE) \\
\midrule
\textbf{Breast Cancer} & Simulated Annealing & 96.48\% $\pm$ 1.76\% \\
(Binary Class.) & ABC Algorithm & \textbf{99.56\% $\pm$ 0.54\%} \\
& LSTM Controller (RL) & \textbf{99.56\% $\pm$ 0.54\%} \\
& Transformer Controller & 98.90\% $\pm$ 0.70\% \\
\bottomrule
\end{tabular}
\end{center}
\end{table}

As detailed in Table I, the Artificial Bee Colony (ABC) algorithm proved robust across all modalities. On the Breast Cancer dataset, both the ABC and the RL controllers quickly reached the dataset's theoretical accuracy ceiling (99.56\%), discovering the absolute optimal architecture for this problem in less than a minute.

However, the regression task revealed a critical vulnerability in standard sequence-generating controllers. While the ABC algorithm reached a highly competitive Negative Mean Squared Error of -0.32 (rivaling manually tuned ensemble methods like Gradient Boosting), the classical LSTM-based RL controller effectively collapsed, returning a sub-optimal negative MSE of -0.39. 

This collapse can be attributed to the nature of the policy gradient updates. In classification, accuracy provides a naturally bounded reward signal $[0, 1]$. In regression, the reward relies on an unbounded negative error metric. Without strict reward scaling, high variance in the regression proxy causes severe instability in the LSTM's policy gradient, preventing it from converging to a stable macro-architecture. Although the Transformer handled this variance slightly better (achieving a peak Negative MSE of -0.31 during its best run, while maintaining an average performance of -0.36 $\pm$ 0.04 across seeds), this experiment confirmed that generative controllers struggle with micro-parameter tuning on continuous regression tasks, heavily justifying the need for a memetic swarm fine-tuning phase.

It is worth acknowledging that on these low-dimensional baseline tasks, the LSTM and Transformer controllers perform comparably, with the LSTM even matching the best classifier on the Breast Cancer dataset. This behavior is consistent with the architectural requirements of these tasks: the optimal topologies are inherently shallow, and the long-range topological dependencies that motivate self-attention only emerge in deeper search spaces. Consequently, these small-scale results do not empirically isolate the Transformer's contribution. A controlled ablation contrasting an LSTM-based warm-start against the Transformer-based warm-start on a long-sequence search space such as CIFAR-10 is left for future work.

\subsection{Implementation Details and Hyperparameters}
For full reproducibility, the core hyperparameters are defined as follows.

\begin{itemize}
    \item \textbf{Transformer controller:} child networks are trained with
    Adam (lr~$=10^{-3}$); the controller uses Adam (lr~$=10^{-2}$) over
    batches of 16 sampled architectures. The depth penalty is $\lambda=0.5$
    and the initial entropy coefficient $\beta=0.05$; upon $N=5$ stagnant
    iterations, $\beta$ is multiplied by $1.5$ (capped at $0.5$). Maximum
    depth is bounded at 20 layers for CIFAR-10 and 50 for the tabular tasks.

    \item \textbf{Metaheuristics:} ABC uses a colony of $P=20$ bees and an
    abandonment threshold \texttt{limit}~$=5$. SA uses $T_0=100$ with a
    cooling rate of $0.99$. The SA neighbourhood and ABC employed/onlooker
    phases share the same \texttt{neighbor} mutation operator.

    \item \textbf{Proxy evaluation:} each candidate is trained for 10 epochs
    (Adam, lr~$=10^{-3}$) on a 40\% training split and ranked on a disjoint
    10\% validation split; the official test set is never used during search.
    The proxy uses accuracy for classification and Average Precision (AUPRC)
    for the imbalanced fraud task. An internal patience of 2 epochs halts
    stagnant trainings.

    \item \textbf{Budget and fairness:} we report the number of proxy
    evaluations (not iterations) as the comparable budget unit. On CIFAR-10,
    the hybrid runs 20 Transformer iterations (batch 16) followed by 15 ABC
    iterations; SA runs 800 iterations and ABC-only 30. To match budgets,
    Random Search is allotted the same number of evaluations as the hybrid.
    The ABC early-stopping patience is 15 within the hybrid and 5 for the
    standalone ABC baseline.

    \item \textbf{Final training:} the selected architecture is trained for
    100 epochs (Adam, lr~$=10^{-3}$, cosine annealing, batch 512) with
    \texttt{RandomCrop(32,\,padding=4)} and \texttt{RandomHorizontalFlip}
    augmentation. For fraud, the decision threshold is selected on validation
    over $[0.05, 0.99]$ and applied once to the test set.

    \item \textbf{Reproducibility:} all metrics are averaged over 3 seeds
    $[42,43,44]$ with sample standard deviation. Experiments use TF32 matmul
    and \texttt{cudnn.benchmark}; reported variance therefore includes both
    seed-to-seed and run-to-run numerical variability on the RTX~3060.
\end{itemize}

\subsection{Memetic Ablation Study and Fair Comparison (CIFAR-10)}
We compared our approach against Random Search, pure Simulated Annealing (SA), and standalone ABC under a strictly matched evaluation budget. As shown in Table \ref{tab:comprehensive_comparison}, hybridization largely outperforms isolated methods. Left to its own devices (Cold-Start), the ABC standalone struggles to navigate the macro-architecture space. Coupled with the Transformer acting as a macro-topological filter, the memetic framework reaches 84.85\% ($\pm$ 2.88\%) in $\sim$3.2 hours of search time. 

Crucially, Table \ref{tab:comprehensive_comparison} demonstrates the effectiveness of our multi-objective reward. When the depth penalty (malus) is removed, the framework achieves a higher accuracy of 86.82\% but at the cost of significant model bloat ($\sim$229k parameters). Activating the malus restricts the capacity to $\sim$174k parameters, proving that the reward function successfully navigates the Pareto frontier of Frugal AI.

As shown in Table \ref{tab:comprehensive_comparison}, we evaluated the algorithms under a strictly matched evaluation budget. While Simulated Annealing (SA) achieves the highest absolute accuracy (87.21\%), it does so by generating massive architectures ($\sim$294k parameters). SA's lack of a structural penalty appears to incentivize capacity expansion rather than targeted feature learning, continuously stacking layers to force accuracy improvements. Consequently, evaluating these oversized models drastically inflates the computational cost, pushing the search time to nearly 2 hours.

Consistent with the literature, Random Search (RS) proves to be a highly robust baseline (83.51\%). However, evaluating RS purely on average performance masks its fundamental lack of reliability. While RS occasionally stumbles upon a compact model (averaging $\sim$133k parameters) purely by stochastic chance, it provides no structural intelligence. As the complexity of the dataset or the search space scales, the probability of RS randomly guessing an optimal topology drops. 

By contrast, our memetic approach provides a policy-driven trajectory. Unlike a stochastic warm-start (e.g., seeding the swarm with the best architecture found via Random Search), the Transformer provides a probabilistically learned structural prior. The 1.34\% accuracy gap achieved by our full-budget Hybrid method over RS is mathematically significant within such a strictly bounded macro-search space. Because the Transformer's policy gradient actively learns the architectural constraints via the Advantage metric, it autonomously directs the swarm away from barren topological plateaus and towards the Pareto frontier.

This behavioral difference is clearly highlighted by the ablation study. Without the depth penalty, the Hybrid method aggressively matches the SA's brute-force performance (86.82\%) but balloons to $\sim$229k parameters. When the depth penalty is activated, the Hybrid method overtakes Random Search (84.85\% vs 83.51\%) while maintaining a highly frugal footprint ($\sim$174k parameters). Furthermore, exploring the extremes of this Pareto frontier reveals that individual runs can achieve great scores: for instance, Seed 43 of the full-budget hybrid reached a peak accuracy of \textbf{88.17\%}, directly rivaling the best Simulated Annealing results but requiring over 130,000 fewer parameters than the SA's equivalent peak. Ultimately, this proves that even under strict operational constraints, the Transformer-guided swarm reliably guarantees highly optimized topologies without relying on luck or endless parameter scaling.

While manual architectures like ResNet-20 ($\sim$270k parameters) achieve $\sim$91.25\% accuracy on CIFAR-10, our framework's objective is strict parameter frugality without human intervention. Our memetic search autonomously discovers architectures that are  1.5 to 5 times smaller depending on the configuration (see Section~\ref{sec:frugality}) than ResNet-20, proving its effectiveness for environments where memory footprint is the primary bottleneck.
It is worth noting the significant discrepancy in the number of evaluations for the standalone ABC algorithm ($\sim$415 evaluations) compared to the allocated budget of the other methods. This lower evaluation count is not an artificial constraint, but a direct consequence of the algorithm repeatedly triggering the early-stopping mechanism (patience) across all random seeds. 

Initialized randomly in a vast, discrete topological space (the "Cold-Start" effect), the standalone bee swarm fundamentally lacks global orientation. Its local mutation operators fail to find a viable gradient of improvement, causing the population to stagnate almost immediately in barren plateaus or poor local minima, which aborts the search. This premature convergence perfectly illustrates why swarm metaheuristics fail in isolation for macro-topological routing, and strongly justifies the necessity of our Transformer's "warm-start" blueprint to guide the swarm toward fruitful regions of the search space.

Our standalone metaheuristic results are consistent with prior swarm-based NAS. Lankford and Grimes \cite{lankford2024} report a mean ACO accuracy of 82.2\% (peak 84.8\%) on CIFAR-10 under a comparable depth-bounded setting, in the same range as our isolated ABC and Random Search baselines. Notably, they also observe that doubling the swarm size yields only a $\sim$1.2\% accuracy gain while roughly doubling run time—mirroring our finding that allocating more budget to an unguided swarm offers diminishing returns and motivating the Transformer warm-start.

\begin{table*}[htbp]
\caption{Comprehensive NAS Performance Comparison on CIFAR-10 (Budget \& Ablation)}
\label{tab:comprehensive_comparison}
\begin{center}
\begin{tabular}{lcccc}
\toprule
\textbf{Algorithm Configuration} & \textbf{Evals (Avg)} & \textbf{Accuracy (Avg $\pm$ Std)} & \textbf{Params (Avg)} & \textbf{Time (min)} \\
\midrule
Random Search & $\sim$770 & 83.51\% $\pm$ 1.32\% & $\sim$133k & $\sim$29 \\
Simulated Annealing & $\sim$800 & \textbf{87.21\% $\pm$ 1.61\%} & $\sim$294k & $\sim$109 \\
ABC Only & $\sim$415 & 76.19\% $\pm$ 1.55\% & $\sim$247k & $\sim$16 \\
\midrule
\textbf{Hybrid (Malus + Early Exit)} & $\sim$770 & 83.39\% $\pm$ 1.75\% & \textbf{$\sim$97k} & $\sim$80 \\
\textbf{Hybrid (Malus + Full Budget)} & $\sim$1000 & 84.85\% $\pm$ 2.88\% & $\sim$174k & $\sim$192 \\
\textbf{Hybrid (No Malus + Full Budget)} & $\sim$1000 & 86.82\% $\pm$ 1.74\% & $\sim$229k & $\sim$233 \\
\bottomrule
\end{tabular}
\end{center}
\end{table*}
\subsection{Comparison with State-of-the-Art NAS Methods}
After validating the core mechanics of our memetic approach, we contextualize \textit{nas-torch} within the broader NAS literature. Table \ref{tab:sota_comparison} summarizes this comparison on the CIFAR-10 dataset.

It is crucial to acknowledge that comparing these methods directly involves fundamentally different search space paradigms. Leading methods like NAS with RL \cite{zoph2017} and DARTS \cite{liu2019} perform a micro-search over highly complex cell structures (e.g., inverted residuals, dilated or separable convolutions) which are then stacked sequentially. While this yields massive theoretical capacity, it drastically inflates both the search space size and the final inference latency on edge devices. Because our primary goal is democratic accessibility and strict frugality, \textit{nas-torch} deliberately restricts its macro-search space to standard operations. 

Therefore, the significantly lower \textbf{Search Cost} of our method is the direct result of an intentional structural trade-off: sacrificing micro-cell complexity to ensure rapid convergence and low-latency topologies on consumer hardware.

\begin{table}[htbp]
\caption{Efficiency and Parameter Comparison with SOTA NAS Methods on CIFAR-10}
\label{tab:sota_comparison}
\begin{center}
\resizebox{\linewidth}{!}{
\begin{tabular}{llccc}
\toprule
\textbf{Method} & \textbf{Search Space} & \textbf{Params} & \textbf{Search Cost} & \textbf{Accuracy} \\
& & (Millions) & (GPU-days) & \\
\midrule
NAS with RL \cite{zoph2017} & Complex Cells & 37.4 & 22,400 & 96.35\% \\
AmoebaNet-A \cite{real2019} & Complex Cells & 3.2 & 3,150 & 96.66\% \\
DARTS \cite{liu2019} & Complex Cells & 3.3 & 1.5 & 97.00\% \\
ENAS \cite{pham2018} & Complex Cells & 4.6 & 0.45 & 97.11\% \\
\midrule
\multicolumn{5}{c}{\textit{*Note: Methods above operate on micro-cell spaces; direct accuracy comparison is not meaningful.}} \\
\midrule
\textbf{Transf. + ABC(no malus+ full budget)} & \textbf{Standard Layers} & \textbf{0.190--0.281} & \textbf{0.16} & \textbf{85\% -- 88\%} \\
\textbf{Simulated Annealing} & \textbf{Standard Layers} & \textbf{0.237--0.360} & \textbf{0.084} & \textbf{86\% -- 89\%} \\
\bottomrule
\end{tabular}%
}
\end{center}
\end{table}
By completing the architectural search in approximately 0.16 GPU-days, our framework is highly resource-efficient. However, as noted in the literature, raw GPU-days can be a misleading metric due to generational hardware improvements. To ensure a fair comparison, we normalize the search cost by the theoretical FP32 compute power of the hardware used (TFLOPS-days). 

Our search on a consumer-grade RTX 3060 ($\sim$13 TFLOPS) yields a normalized cost of roughly \textbf{2.08 TFLOPS-days}. In contrast, the foundational NAS with RL \cite{zoph2017} utilizing K40 GPUs ($\sim$4.3 TFLOPS) required over 96,000 TFLOPS-days, while more efficient gradient-based methods like DARTS \cite{liu2019} on GTX 1080 Ti ($\sim$11.3 TFLOPS) require approximately 17 TFLOPS-days. While this theoretical peak FP32 normalization does not account for memory-bound operational bottlenecks or effective hardware utilization rates, it provides a strict lower-bound estimation demonstrating that our memetic approach reduces the computational overhead by roughly 6–7× lower than DARTS, while discovering highly compact feature extractors suitable for accessible AI development.
Beyond gradient-based methods, swarm-based NAS also incurs substantial wall-clock cost: on CIFAR-10, the PSO and ACO searches of Lankford and Grimes \cite{lankford2024} required between 9 (ACO) and 22 (PSO) hours on a GTX~1080~Ti, with run time scaling sharply as the swarm size grows. Our memetic search  completes in at most a few hours on more modest hardware, underscoring the practical accessibility of the framework.

\subsection{Architectural Frugality and Parameter Efficiency}
\label{sec:frugality}
The impact of our memetic framework lies in its inherent parameter efficiency. For context, standard architectures commonly deployed on CIFAR-10, such as ResNet-18 or MobileNetV2, contain approximately 11 million and 3.4 million parameters, respectively.

By integrating a length penalty directly into the Transformer's reward function, our framework actively selects against model bloat. The final architectures discovered by our optimizers are compact. For instance, an early-stopping variant of the hybrid model generated by Seed 42 achieved 81.47\% accuracy with only \textbf{50,650 parameters}, and Seed 43 reached 83.82\% with \textbf{77,002 parameters}.

While state-of-the-art hand-crafted models like ResNet-20 achieve higher absolute accuracy ($\sim$91.25\%) with approximately 270,000 parameters, our results should be interpreted as a proof of concept for fully autonomous, resource-constrained NAS. A common alternative for edge deployment is applying network pruning or quantization to such oversized models. However, pruning inherently relies on human expert design to establish the initial macro-architecture and requires a massive computational budget to pre-train the dense network before compression. 

In contrast, our framework operates with zero human bias. It autonomously discovers functional architectures that are 1.5× to 5× smaller than ResNet-20 directly from scratch. Although this yields a lower absolute accuracy on CIFAR-10, it successfully demonstrates that highly efficient, task-specific feature extractors can be generated entirely automatically on a standard 6GB consumer GPU. This effectively democratizes the architectural design process, allowing researchers and engineers to bypass heavy pre-training phases and human empirical tuning entirely.

\subsection{Flexibility \& Imbalanced Data (Credit Card Fraud)}
To demonstrate the agnosticism of our framework, we tested it on a massively asymmetric tabular bank fraud dataset (where frauds represent merely 0.2\% of the $\sim$284,000 total transactions). The reward function was rewritten to optimize the F1-Score rather than pure Accuracy. Entirely autonomously, the framework converged to an architecture reaching an F1-Score of 0.7178 $\pm$ 0.0279 (Recall: 0.8095, Precision: 0.6467) and an AUPRC of 0.7698. While heavily tuned gradient boosting methods can achieve higher absolute scores on this dataset, our NAS framework automatically designed a neural feature extractor containing only \textbf{4,614 parameters}. This highlights the framework's ability to adapt to non-vision modalities and optimize threshold-independent metrics without human intervention.

\subsection{Search Space Capacity Bottleneck (CIFAR-100)}
To evaluate the scalability of our restricted search space, we conducted a memetic search on the more complex CIFAR-100 dataset. For this stress test, the optimizers utilized the full evaluation budget (without early stopping) while strictly retaining the depth penalty ($\lambda = 0.5$). The results across three random seeds are detailed in Table \ref{tab:cifar100}.

\begin{table}[htbp]
\caption{Hybrid NAS Performance on CIFAR-100 (100 Epochs)}
\label{tab:cifar100}
\begin{center}
\begin{tabular}{lccc}
\toprule
\textbf{Seed} & \textbf{Accuracy} & \textbf{Params} & \textbf{Search Time} \\
\midrule
42 & 57.58\% & $\sim$449k & $\sim$49 min \\
43 & 52.71\% & $\sim$232k & $\sim$53 min \\
44 & 46.79\% & $\sim$3.10M & $\sim$67 min \\
\midrule
\textbf{Mean} & \textbf{52.36\% $\pm$ 5.40\%} & \textbf{$\sim$1.26M} & \textbf{$\sim$56 min} \\
\bottomrule
\end{tabular}
\end{center}
\end{table}

The average accuracy of 52.36\% ($\pm$ 5.40\%) indicates a clear mathematical limit of our restricted macro-vocabulary. While the framework successfully extracts features up to a certain point (reaching 57.58\% on Seed 42), resolving a 100-class problem typically requires advanced macro-cells (e.g., Inverted Residuals, Dense blocks) rather than simple standard convolutions. 

Furthermore, this experiment exposes the vulnerability of our highly frugal proxy evaluation split. On CIFAR-100, allocating 10\% of the data to validation leaves only $\sim$50 images per class. This extremely low sample density injects severe noise into the reward signal during the 10-epoch proxy training. Crucially, as observed in Seed 44 (which exploded to $\sim$3.10M parameters), this proxy noise can become so erratic that it occasionally overwhelms the regularizing effect of the depth penalty. Forced to complete its full evaluation budget without early stopping, the controller stacked layers uncontrollably in a failing attempt to fit the unstable validation signal. This confirms that while our framework is highly efficient for targeted tasks and edge deployment, scaling to massive multi-class datasets will require both enriching the search space and stabilizing the proxy sampling strategy.

\section{Conclusion}
This paper presented a memetic NAS framework validating the hypothesis that it is possible to automatically generate high-performing and tailored networks without requiring supercomputers. By coupling the generative intelligence of a dynamic entropy autoregressive Transformer with the granular optimization of a bee swarm, the architecture search gains in stability, efficiency, and frugality. 

Future work will proceed along three primary axes. First, to overcome the capacity bottleneck identified on complex datasets like CIFAR-100, we plan to enrich our search space vocabulary with more advanced, modern macro-cells (such as Inverted Residuals or Dense Blocks). The modular nature of our \textit{layer\_classes} ensures that the Transformer will be able to orchestrate these complex blocks without requiring changes to the underlying search algorithm. Second, we aim to integrate training-free evaluation metrics (Zero-Cost Proxies, such as SynFlow or Fisher Information \cite{abdelfattah2021}) to further reduce the proxy evaluation time from minutes to mere milliseconds. Finally, we will focus on establishing a strict hardware-aware Pareto front by incorporating direct latency and memory footprint measurements into the multi-objective reward function, fully realizing the vision of surgical, edge-ready AI deployment.

\section*{Acknowledgment}
The author would like to thank Florentin THULLIER for his guidance throughout this project, and Kevin BOUCHARD for his valuable feedback and proofreading.


\begin{thebibliography}{00}
\bibitem{zoph2017} B. Zoph and Q. V. Le, "Neural Architecture Search with Reinforcement Learning," in \textit{Proc. of ICLR}, 2017.
\bibitem{pham2018} H. Pham, M. Y. Guan, B. Zoph, Q. V. Le, and J. Dean, "Efficient Neural Architecture Search via Parameter Sharing," in \textit{Proc. of ICML}, 2018.
\bibitem{liu2019} H. Liu, K. Simonyan, and Y. Yang, "DARTS: Differentiable Architecture Search," in \textit{Proc. of ICLR}, 2019.
\bibitem{abdelfattah2021} M. S. Abdelfattah, A. Mehrotra, Ł. Dudziak, and N. D. Lane, "Zero-Cost Proxies for Lightweight NAS," in \textit{Proc. of ICLR}, 2021.
\bibitem{yu2025} C. Yu, X. Liu, Y. Wang, Y. Liu, W. Feng, D. Xiong, C. Tang, and J. Lv, "GPT-NAS: Evolutionary Neural Architecture Search with the Generative Pre-Trained Model," \textit{arXiv preprint arXiv:2305.05351}, 2025.
\bibitem{real2019} E. Real, A. Aggarwal, Y. Huang, and Q. V. Le, "Regularized Evolution for Image Classifier Architecture Search," in \textit{Proc. of AAAI}, 2019.
\bibitem{lankford2024} S. Lankford and D. Grimes, "Neural Architecture Search using Particle Swarm and Ant Colony Optimization," \textit{arXiv preprint arXiv:2403.03781}, 2024.
\bibitem{li2019random} L. Li and A. Talwalkar, "Random Search and Reproducibility for Neural Architecture Search," in \textit{Proc. of UAI}, 2019.
\end{thebibliography}
\end{document}